\documentclass[conference]{IEEEtran}

\usepackage{amsthm}

\usepackage{lineno}
\usepackage{amsfonts}
\usepackage{amssymb}
\usepackage{booktabs}
\usepackage[cmex10]{amsmath}
\usepackage{algorithm}
\usepackage[noend]{algpseudocode}
\usepackage{multirow}
\usepackage{color}
\usepackage{graphicx}
\usepackage{svg}
\usepackage{amsfonts}
\usepackage{amssymb}
\usepackage{booktabs}
\usepackage[cmex10]{amsmath}
\usepackage{algorithm}
\usepackage[noend]{algpseudocode}
\usepackage{authblk}
\usepackage{hyperref}

\newcommand{\revise}[1]{#1}

\begin{document}


\title{Visual Style Prompt Learning  Using Diffusion Models for Blind Face Restoration}

\author[1,2]{Wanglong Lu}
\author[1]{Jikai Wang}
\author[3]{Tao Wang}
\author[4]{Kaihao Zhang}
\author[2]{Xianta Jiang}
\author[1]{Hanli Zhao$^{*}$}

\affil[1]{Key Laboratory of Intelligent Informatics for Safety and Emergency of Zhejiang Province, Wenzhou University, China}
\affil[2]{Department of Computer Science, Memorial University of Newfoundland, St. John's, Canada}
\affil[3]{State Key Laboratory for Novel Software Technology, Nanjing University, China}
\affil[4]{College of Engineering and Computer Science, Australian National University, Australia}
\affil[$^{*}$]{Corresponding author: hanlizhao@wzu.edu.cn}

\maketitle

\begin{abstract}
Blind face restoration aims to recover high-quality facial images from various unidentified sources of degradation, posing significant challenges due to the minimal information retrievable from the degraded images. 
Prior knowledge-based methods, leveraging geometric priors and facial features, have led to advancements in face restoration but often fall short of capturing fine details.
To address this, we introduce a visual style prompt learning framework that utilizes diffusion probabilistic models to explicitly generate visual prompts within the latent space of pre-trained generative models. These prompts are designed to guide the restoration process.
To fully utilize the visual prompts and enhance the extraction of informative and rich patterns, we introduce a style-modulated aggregation transformation layer.
Extensive experiments and applications demonstrate the superiority of our method in achieving high-quality blind face restoration. The source code is available at \href{https://github.com/LonglongaaaGo/VSPBFR}{https://github.com/LonglongaaaGo/VSPBFR}.
\end{abstract}






\begin{IEEEkeywords}
Denoising diffusion probabilistic models, generative adversarial networks, blind face restoration.
\end{IEEEkeywords}








\section{Introduction}

Blind face restoration is dedicated to reconstructing high-fidelity facial images from a range of unidentified sources of degradation such as blurring, noise, downsampling, and compression artifacts~\cite{wang2022survey,HSIEH2021107597_pr,SHEN2022108909_pr}. Given the complexity and un-retractability of the degradation in real-world situations, only limited information could be utilized from the compromised face images~\cite{WANG2023109360_pr,ZHAO2022108983_pr}. Reconstructing facial images with faithful features from blindly degraded images stands as a pivotal challenge in computer vision and image processing~\cite{WANG2024109956_pr,WANG2022102321,lu2023grig,TensorSVD}.


Currently, prior knowledge-based methods are leading the way in achieving high-quality face restoration. 
These methods use geometric priors to guide the restoration process, by drawing on shape information such as parsing maps~\cite{Chen_2021_CVPR}. Nonetheless, their results often lack comprehensive detail (like identity information) since they focus on limited facial features~\cite{Wang_2022_CVPR}.
Some studies~\cite{Wang_2021_CVPR,Menon_2020_CVPR,Wang_2022_CVPR,wang2023restoreformer++} further employed facial priors inherently present in pre-trained generative models~\cite{Karras2019} or vector quantization dictionary~\cite{VQVAE} for guiding restorations that retain detailed facial attributes.
However, when facial images are degraded, critical features (e.g., details of expression and identity) may be lost, leading to inaccurate facial prior estimation for reconstruction. Achieving precise facial prior estimation that corresponds to high-quality images is crucial for enhanced restoration, and is still an ongoing challenge.






Recent studies have shown that using embeddings of pre-trained models as visual prompts~\cite{potlapalli2023promptir,Xia_2023_ICCV,wang2024promptrr} enhances restoration processes, highlighting the importance of visual priors for restoration quality. Yet, these prompts lack easy visualization and interpretation, which constrains the understanding of their impact on performance.
In contrast, generative models provide dense latent representations that encapsulate visual styles and attributes, offering valuable guidance.
By leveraging generative models' latent representations, we can introduce a more intuitive and reliable means to direct the face restoration process. Nonetheless, research in this area remains limited.

In this paper, we explore the use of latent representations in pre-trained generative models as visual style prompts for directing face restoration. 
We plan to explore (1) how to encode a degraded face image into high-quality visual prompts matching ground-truth images, and (2) how to integrate the visual prompts and facial priors within the network to boost performance. 



To tackle these challenges, we first explore the role of Diffusion Probabilistic Models (DMs)~\cite{NEURIPS2020DDPM} in improving the estimation of clean (high-quality) latent representations from degraded images, aiding facial feature extraction for restoration. 
GAN inversion methods can effectively embed a given high-quality image into the corresponding latent space. However, the optimization-based~\cite{Pivotal_tuning2022} is time-consuming, and the encoder-based methods~\cite{Tov2021} offer faster but sometimes less precise mapping. 
Recent studies on DMs show their potential in learning distributions effectively. Yet, their use in embedding latent representations is still not widely explored. Also, the application of these models in processing degraded face images remains insufficiently studied.


Secondly, we evaluate how rescaling and adjusting multiscale convolutional kernels based on visual prompts can enhance feature extraction, and optimize the use of visual information across different receptive fields.
It is crucial for the network to adeptly utilize valuable guidance to improve restoration. Methods like spatial feature transform~\cite{Wang_2021_CVPR} and deformable operations~\cite{gu2022vqfr} yield promising feature extraction but are limited by the narrow receptive fields of convolutional layers, restricting global context usage. 
While Transformer-based methods~\cite{Wang_2022_CVPR,wang2023restoreformer++} excel in face restoration, they mainly require substantial GPU memory and computational resources. Moreover, these methods did not explore the usage of visual prompts.






We thus propose a novel visual style prompt learning framework for blind face restoration.
Thanks to the well-disentangled and capable StyleGAN~\cite{Karras2020} latent space in representing diverse facial attributes, we leverage the latent representations of a pre-trained StyleGAN model as visual prompts to generate potential facial features and inform the subsequent restoration process.
To efficiently transform a degraded face image into high-quality latent representations, a diffusion-based style prompt module progressively performs denoising steps, refining low-quality visual prompts into higher-quality (clean) counterparts. 
The resulting candidate facial features and clean visual prompts are then employed to guide the restoration process within the restoration auto-encoder. 
Furthermore, we introduce a Style-Modulated AggRegation Transformation (SMART) layer to fully utilize visual prompts and capture both contextual information and detailed patterns for better context reasoning.
Extensive comparisons and analysis compared to the state-of-the-art (SOTA) methods on four public datasets demonstrate the effectiveness of our approach in achieving high-quality blind face restoration. We also extend our method to applications like facial landmark detection and emotion recognition, demonstrating its effectiveness in various face-related tasks.



In summary, our paper offers four key contributions:
\begin{itemize}
\item We introduce a diffusion-based style prompt method that uses DMs to generate visual prompts from degraded images in a generative model's latent space, offering clear style cues for effective restoration.


\item We proposed a style-modulated aggregation transformation, which effectively leverages visual prompts and excels in capturing comprehensive contextual information and patterns for boosting performance.

\item We present a visual style prompt learning framework for high-quality blind face restoration that leverages style prompts to guide facial attribute restoration and uses style-modulated aggregation transformation layers to improve clarity and details.


\item We conducted various experiments to demonstrate that our method not only excels in enhancing the quality of images in both synthesized and real-world blind face restoration datasets but also benefits various applications, highlighting the effectiveness of our proposed components. 
\end{itemize}

    

    
 
 


\begin{figure*}[t]
	\includegraphics[width=\textwidth]{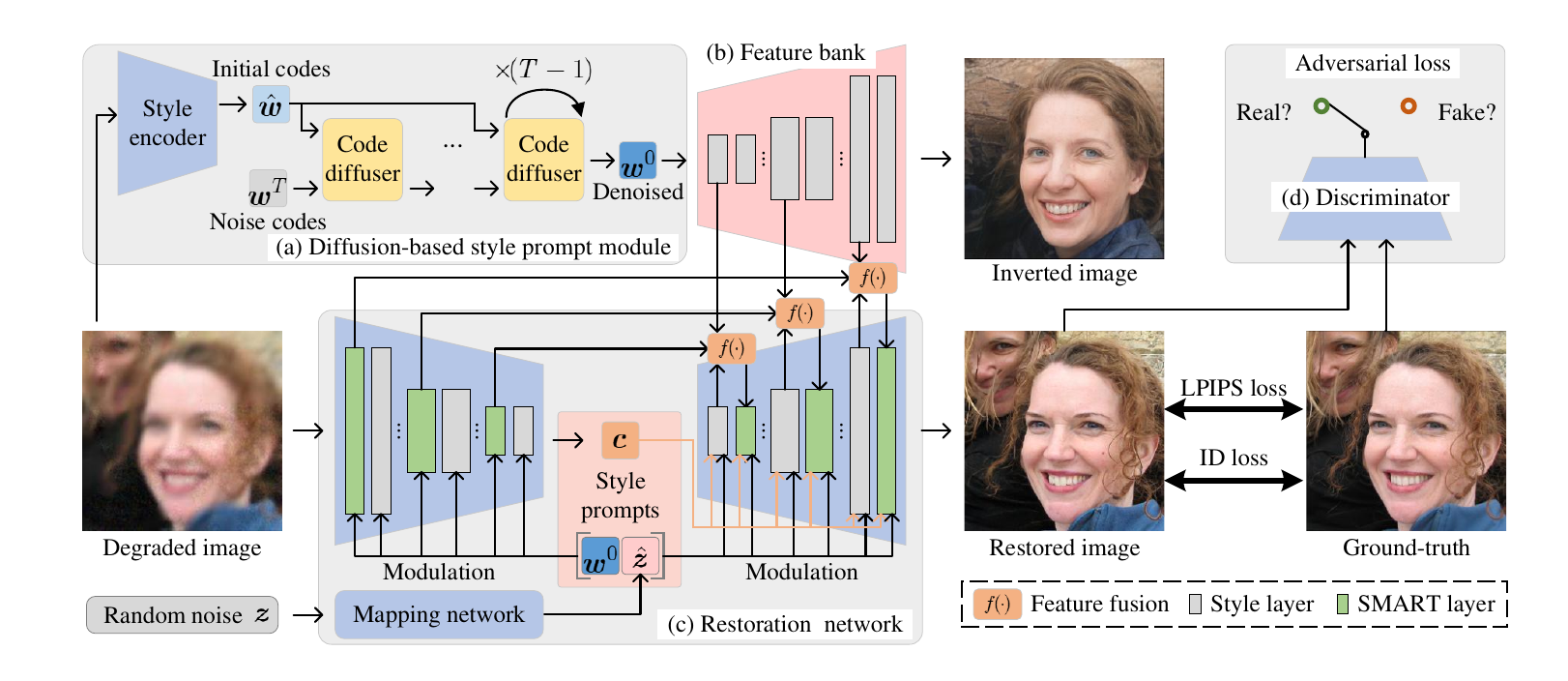}
     \caption{
    The overall pipeline of our framework: the degraded image is processed through a diffusion-based style prompt module (a) to get denoised codes $\boldsymbol{w}^0$ through $T$ diffusion steps, beginning from noise codes $\boldsymbol{w}^T$.
    Then, the restoration auto-encoder (c) processes the degraded image, using the denoised codes $\boldsymbol{w}^0$, random codes $\hat{\boldsymbol{z}}$, and a global code $\boldsymbol{c}$ as style prompts. The network also leverages prior features from the facial feature bank (b), integrating them through a fusion process $f(\cdot)$, to achieve the restored image. 
    } 
	\label{fig:fig_framework}
\end{figure*}


\section{Related work}\label{sec:relatedwork}

\subsection{Blind face restoration}
\revise{Blind face restoration~\cite{ZHANG2024127271}, which aims to reconstruct original face images from degraded ones with unknown sources of degradation, has been a long-standing problem with wide-ranging application potentials, such as real-world video face restoration~\cite{Chen_2024_CVPR}.}
Building on the success of Deep Neural Networks (DNN) in various fields and their strong feature extraction capabilities, DNN-based methods have emerged as the leading approach in blind face restoration~\cite{Chen_2024_CVPR}.
Since limited restoration cues are available for effective restoration, the auxiliary priors have been commonly employed in DNN-based methods. These priors include geometric priors~\cite{Chen_2021_CVPR}, generative priors~\cite{Menon_2020_CVPR,Wan_2020_CVPR,Wang_2021_CVPR,Yang_2021_CVPR}, and multi-priors~\cite{Zhu_2022_CVPR}. 
For geometric priors, an auxiliary model is employed to predict facial structure information, such as landmarks~\cite{Chen_2018_CVPR} and parsing maps~\cite{Chen_2021_CVPR}. However, the accuracy of geometric priors is heavily dependent on the condition of the facial degradation.
Generative priors~\cite{Menon_2020_CVPR,Wan_2020_CVPR,Wang_2021_CVPR,Yang_2021_CVPR}, leveraging high-quality face generators or vector quantization dictionaries~\cite{VQVAE, Wang_2022_CVPR, wang2023restoreformer++}, have shown significant potential in blind face restoration.
However, in scenarios where facial images are degraded, facial attributes like eye details and nasal structure become obscured, resulting in imprecise estimation of facial priors for reconstruction. 
Moreover, they also face challenges in dynamically selecting relevant features and adjusting the kernels' receptive fields.
Recent studies~\cite{yang2023pgdiff} have employed diffusion models for boosting robustness against common degradations, successfully generating high-quality facial images. Yet, these approaches, mainly operating in the pixel space, suffer from low inference speeds and are time-intensive.
In contrast, we utilize a denoising process in the latent space to restore clear representations of degraded images with high efficiency. Our model combines dynamically rescaling and adjusting convolutional kernels to enhance the extraction of informative features and detailed patterns for better restoration.

\subsection{Generative image synthesis}
Generative Adversarial Networks (GANs) have excelled in generating high-quality images, with models like StyleGANs~\cite{Karras2019,Karras2020} enabling significant advancements in latent code inversion and editing for tasks like image restoration~\cite{Wang_2021_CVPR} and manipulation~\cite{Tov2021,lu2022inpainting,FACEMUG}. Recently, Diffusion Probabilistic Models (DMs)\cite{NEURIPS2020DDPM} have further improved image synthesis quality through parameterized Markov chains. While some studies have begun exploring DMs for image restoration\cite{Qiu2023DiffBFRBD,DifFace}, our approach focuses on leveraging diffusion and denoising processes within the style latent space~\cite{Karras2020} to predict visual style prompts that guide blind face restoration.

\section{Method}\label{sec:algo}

\subsection{Overview}

 As shown in Fig.~\ref{fig:fig_framework}, given a degraded facial image $\mathbf{I}_{de} \in \mathbb{R}^{h \times w \times 3}$, our framework aims to restore the degraded facial image to achieve both visual and structural fidelity compared to the ground-truth image $\mathbf{I}_{gt} \in \mathbb{R}^{h \times w \times 3}$. We define $\mathcal{W}+$ as the style latent space~\cite{Karras2020}.



\textbf{Diffusion-based style prompt module.} As shown in Fig.~\ref{fig:fig_framework} (a), our style prompt module predicts denoised latent codes from a degraded facial image. First, a style encoder $E_{\theta_{e}}(\cdot)$ with the network parameters $\theta_{e}$, embeds the given degraded image to initial latent codes $\hat{\boldsymbol{w}} \in \mathbb{R}^{512 \times N} = E_{\theta_{e}}(\mathbf{I}_{de}) \in \mathcal{W}+$. Guided by the $\hat{\boldsymbol{w}}$, our code diffuser $P_{\theta_{p}}(\cdot)$ (with the network parameters $\theta_{p}$) samples Gaussian random noise codes $\boldsymbol{w}^T \in \mathbb{R}^{512 \times N}$ and then gradually denoise $\boldsymbol{w}^T$ with $T$ steps to get denoised latent codes $\boldsymbol{w}^0 \in \mathbb{R}^{512 \times N} \in \mathcal{W}+$. For each step $t$, code diffuser gradually predict the noise: $\hat{\boldsymbol{\epsilon}}^{t} \in \mathbb{R}^{512 \times N} = P_{\theta_{p}}(\boldsymbol{w}^t,\hat{\boldsymbol{w}},t)$. The denoising step is performed by subtracting the predicted noise from $\boldsymbol{w}^t \in \mathbb{R}^{512 \times N}$. $N$ is the number of style vectors in latent codes.

\textbf{Facial feature bank.} 
As shown in Fig.~\ref{fig:fig_framework} (b), we leverage the StyleGAN generator $S_{\theta_{s}}$ as our facial feature bank.  The generator uses denoised latent codes $\boldsymbol{w}^0$ to produce multi-scale, coarse facial features.  From $S_{\theta_{s}}$, we can obtain a set of facial feature maps $\mathcal{F}^s =  \{\mathbf{F}^s_i \in \mathbb{R}^{\hat{h}_{i} \times \hat{w}_{i} \times \hat{c}_{i}} | i \in [1,N]\}$ and an inverted image $\mathbf{I}_{ve} \in \mathbb{R}^{h \times w \times 3}$, such that  $(\mathcal{F}^s,\mathbf{I}_{ve}) = S_{\theta_{s}}(\boldsymbol{w}^0)$. The dimensions $\hat{h}_i \times \hat{w}_i \times \hat{c}_i$ represent the size of the feature maps at the $i$-th layer.

\textbf{Restoration auto-encoder.} As shown in Fig.~\ref{fig:fig_framework} (c), 
a multi-layer fully-connected neural mapping network $F_{\theta_{f}}$ (with the network parameters $\theta_{f}$) generates a random noise vector $\boldsymbol{z} \in \mathbb{R}^{512 \times 1}$ as the input to get random styles $\hat{\boldsymbol{z}} \in \mathbb{R}^{512 \times 1} = F_{\theta_{f}}(\boldsymbol{z})$.
The restoration auto-encoder $G_{\theta_g}$ leverages $\mathbf{I}_{de}$, $\boldsymbol{w}^0$, $\hat{\boldsymbol{z}}$, and $\mathcal{F}^s$ to generate a restored image $\mathbf{I}_{out}= G_{\theta_g}(\mathbf{I}_{de},\boldsymbol{w}^0,\hat{\boldsymbol{z}},\mathcal{F}^s) \in \mathbb{R}^{h \times w \times 3}$. 

\textbf{Discriminator.} 
A discriminator network $D_{\theta_{d}}$ with trainable parameters $\theta_{d}$, is trained to distinguish real images from the generated ones (\textit{e.g.}, $\mathbf{I}_{out}$ or $\mathbf{I}_{gt}$). For a given image $\mathbf{I}$, we have  $D_{\theta_{d}}(\mathbf{I}) \in \mathbb{R}^{1 \times 1}$.

\begin{figure*}[t]
    \centering
	\includegraphics[width=0.80\textwidth]{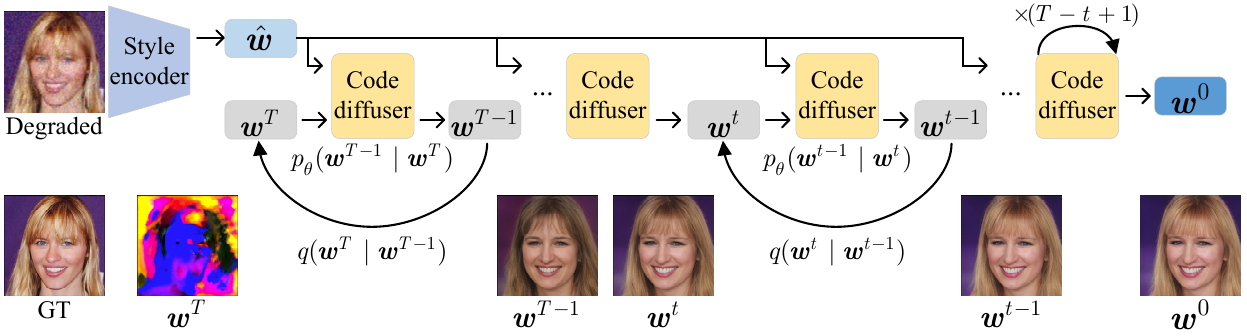}
     \caption{
    The detailed diffusion ($\leftarrow$) and denoising ($\rightarrow$) processes in the style latent space. We also show the corresponding inverted images of latent codes in steps.
    } 
	\label{fig:fig_diffusion}
\end{figure*}


\subsection{Diffusion-based style prompt module}\label{sec:style_prompts}

\revise{Our diffusion-based style prompt module involves a two-stage refinement process. It uses the style encoder to predict initial guidance and then applies a denoising process to sample high-quality latent codes. These refined latent codes serve as visual prompts for generating candidate facial features.}

 Guided by the initial codes $\hat{\boldsymbol{w}}$, the code diffuser leverages diffusion models (DMs)~\cite{NEURIPS2020DDPM} to produce a denoised one. DMs have a diffusion process and a denoising process. The former is for training, and the latter is to sample latent codes from random Gaussian noise. As shown in Fig.~\ref{fig:fig_diffusion}, the diffusion process (a.k.a. forward process) incrementally introduces Gaussian noise into the data. This process transforms the clean latent codes $\boldsymbol{w}^0$ into an approximately pure Gaussian noise $\boldsymbol{w}^T$ using a variance schedule $\beta_1, \ldots, \beta_T$. The diffusion process is defined as:
\begin{equation}\label{equ:code_diffusion}
	\begin{aligned}
        q(\boldsymbol{w}^{t}|\boldsymbol{w}^{t-1}) = \mathcal{N}(\boldsymbol{w}^{t}; \sqrt{1 - \beta_{t}}\boldsymbol{w}^{t-1}, \beta_{t}\mathbf{I}).
    \end{aligned}
\end{equation}
where $\boldsymbol{w}^t$ can be directly approximated by $\boldsymbol{w}^t=\sqrt{\bar{\alpha}_t} \boldsymbol{w}^0+ \sqrt{1-\bar{\alpha}_t} \boldsymbol{\epsilon}$, with $\bar{\alpha}_t=\prod_{s=1}^t \alpha_s$, $\alpha_t=1-\beta_t$, and $\boldsymbol{\epsilon} \sim \mathcal{N}(\mathbf{0}, \mathbf{I})$. 

As shown in Fig.~\ref{fig:fig_diffusion}, the denoising process is designed to sample the cleaner version $\boldsymbol{w}^{t-1}$ from $\boldsymbol{w}^{t}$ by estimating the added noise, which is defined as:
\begin{equation}\label{equ:code_denoise}
	\begin{aligned}
        p_{\theta}(\boldsymbol{w}^{t-1} | \boldsymbol{w}^t) = \mathcal{N}(\boldsymbol{w}^{t-1}; \boldsymbol{\mu}_{\theta}(\boldsymbol{w}^t, t), \boldsymbol{\Sigma}_{\theta}(\boldsymbol{w}^t, t)).
    \end{aligned}
\end{equation}
To sample the denoised latent codes $\boldsymbol{w}^0$ by using our code diffuser, we iteratively denoise $\boldsymbol{w}^t$ from $t=T$ to $t=1$. The Eq.~\ref{equ:code_denoise} is implemented as:
\begin{equation}\label{equ:im_code_denoise}
	\begin{aligned}
        \boldsymbol{w}^{t-1}=\frac{1}{\sqrt{\alpha_t}}\left(\boldsymbol{w}^t-\frac{1-\alpha_t}{\sqrt{1-\bar{\alpha}_t}} \hat{\boldsymbol{\epsilon}}^{t} \right)+\sigma_t \overline{\boldsymbol{\epsilon}},
    \end{aligned}
\end{equation}
where  $\hat{\boldsymbol{\epsilon}}^{t}= P_{\theta_{p}}(\boldsymbol{w}^t,\hat{\boldsymbol{w}},t)$; variance $\sigma_t^2=\frac{1-\bar{\alpha}_{t-1}}{1-\bar{\alpha}_t} \beta_t$ and noise $\overline{\boldsymbol{\epsilon}} \sim \mathcal{N}(\mathbf{0}, \mathbf{I})$. 



\textbf{Code diffuser.}
\revise{Our code diffuser is with four temporal-aware code-to-code blocks (TACC). 
Each block is an extension of FFCLIP's semantic modulation block~\cite{zhu2022one}. It sets $\Gamma^t_{i-1}$ ($\Gamma^t_{0} = \boldsymbol{w}^{t}$), initial codes $\hat{\boldsymbol{w}}$, and denoising step $t$ as inputs, and outputs intermediate results: $\Gamma^t_{i} = \operatorname{TACC}(\Gamma^t_{i-1},\hat{\boldsymbol{w}},t), i \in [1, 2, 3, 4]$, and $\hat{\boldsymbol{\epsilon}}^{t} = \Gamma^t_{4}$.
Given an intermediate input $\Gamma^t_{i-1}$, each TACC block first calculates query, key, and value embeddings as:
\begin{eqnarray}\label{equ:latent_warping_module1}
	\begin{aligned}
		&\boldsymbol{w}^q =\operatorname{FC}([\hat{\boldsymbol{w}},t]), \boldsymbol{w}^k=\operatorname{FC}(\Gamma^t_{i-1}), \boldsymbol{w}^v=\operatorname{FC}(\Gamma^t_{i-1}), \\
    &\bar{\boldsymbol{w}}^q =\operatorname{FC}([\hat{\boldsymbol{w}},t]), \bar{\boldsymbol{w}}^k=\operatorname{FC}(\Gamma^t_{i-1}), \bar{\boldsymbol{w}}^v=\operatorname{FC}(\Gamma^t_{i-1}), \\
	\end{aligned}
\end{eqnarray}
Then, the channel-based and position-based attention~\cite{fu2019dual}, as well as gated maps and bias are computed to get the intermediate output $\Gamma^t_{i}$:
\begin{eqnarray}\label{equ:latent_warping_module2}
\resizebox{.96\linewidth}{!}{$
	\begin{aligned}
        &\boldsymbol{a} =  {\boldsymbol{w}}^v \cdot \operatorname{Softmax}( {\boldsymbol{w}}^{q^{\top}} \cdot  {\boldsymbol{w}}^k  / \kappa_1) + \operatorname{Softmax}(\bar{\boldsymbol{w}}^q \cdot \bar{\boldsymbol{w}}^{k^{\top}}  / \kappa_2) \cdot \bar{\boldsymbol{w}}^v ,\\
		&\Gamma^t_{i} = \operatorname{LayerNorm}(\boldsymbol{a}) \odot (\sigma(\operatorname{MLP}([\hat{\boldsymbol{w}},t])) + \mathbf{1}) + \phi(\operatorname{MLP}([\hat{\boldsymbol{w}},t])),
	\end{aligned}
    $}
\end{eqnarray}
where $\boldsymbol{w}^q$, $\boldsymbol{w}^k$, $\boldsymbol{w}^v$, $\bar{\boldsymbol{w}}^q$, $\bar{\boldsymbol{w}}^k$, $\bar{\boldsymbol{w}}^v$, $\boldsymbol{a}$, and $\Gamma^t_{i}$ share the same dimension as $\hat{\boldsymbol{w}}$;
$\kappa_1=\sqrt{N}$; $\kappa_2=\sqrt{512}$;
$[\cdot]$ is the concatenation operation; 
$\operatorname{FC}(\cdot)$ is a fully connected layer;  $\operatorname{MLP}(\cdot)$ refers to a stack of two fully connected layers; $\operatorname{Softmax(\cdot)}$ is the softmax activation; $\sigma(\cdot)$ denotes the sigmoid activation; $\phi(\cdot)$ corresponds to the LeakyReLU activation with the negative slope of $0.2$;
$\operatorname{LayerNorm}(\cdot)$ is the LayerNorm layer.}

\subsection{Restoration auto-encoder}\label{sec:restore_net}
\revise{Our restoration auto-encoder utilizes denoised latent codes, {random styles}, and facial priors to recover the degraded image $\mathbf{I}_{de}$. We utilize random styles~\cite{Zhao2021} to improve the diversity.
We also introduce the SMART layer to capture contextual information and detailed patterns fully to enhance context reasoning while using latent codes to guide the restoration process.}

As shown in Fig.~\ref{fig:fig_framework} (c), given denoised latent codes $\boldsymbol{w}^{0}$, random styles $\hat{\boldsymbol{z}}$, and facial priors $\mathcal{F}^s =  \{\mathbf{F}^s_i \in \mathbb{R}^{\hat{h}_{i} \times \hat{w}_{i} \times \hat{c}_{i}} | i \in [1,N]\}$, the features $\mathcal{F}^{en} =  \{\mathbf{F}^{en}_i \in \mathbb{R}^{\hat{h}_{i} \times \hat{w}_{i} \times \hat{c}_{i}} | i \in [1,N]\}$ in our restoration encoder are defined as: 
\begin{equation}\label{equ:decoder}
	\begin{aligned}
		\mathbf{F}^{en}_{i} = \begin{cases} \operatorname{SMART}(\mathbf{F}^{en}_{i-1},[\boldsymbol{w}^0_{i},\hat{\boldsymbol{z}}]), & \text {if } i \bmod 2 = 1;\\
        (\operatorname{SC}(\mathbf{F}^{en}_{i-1},[\boldsymbol{w}^0_{i},\hat{\boldsymbol{z}}]))\downarrow_{2}, & \text {otherwise}, \end{cases}
	\end{aligned}
\end{equation}
where $\operatorname{SC}(\cdot)$ is the style layer; $\mathbf{F}^{en}_{0}=\operatorname{Conv}(\mathbf{I}_{de})$. When $i=N$, we can further get the global code $\boldsymbol{c} = \operatorname{FC}(\mathbf{F}^{en}_{N})$.
Our restoration decoder then generate features $\mathcal{F}^{de} =  \{\mathbf{F}^{de}_i \in \mathbb{R}^{\hat{h}_{i} \times \hat{w}_{i} \times \hat{c}_{i}} | i \in [1,N]\}$, which can be defined as follows: 
\begin{equation}\label{equ:gen_f_de}
	\begin{aligned}
		\mathbf{F}^{de}_{i} = \begin{cases} \operatorname{SC}((\mathbf{F}^{de}_{i-1})\uparrow_{2},[\boldsymbol{w}^0_{i},\hat{\boldsymbol{z}},\boldsymbol{c}]), & \text {if } i \bmod 2 = 1;\\
			\operatorname{SMART}(\hat{\mathbf{F}}^{de}_{i-1},[\boldsymbol{w}^0_{i},\hat{\boldsymbol{z}},\boldsymbol{c}]), & \text {otherwise}, \end{cases}
	\end{aligned}
\end{equation}
where $\mathbf{F}^{de}_{0}=\operatorname{Conv}(\mathbf{F}^{en}_{N})$; $\hat{\mathbf{F}}^{de}_{i-1} = f(\mathbf{F}^{de}_{i-1},{\mathbf{F}}^{en}_{N-i+1},{\mathbf{F}}^{s}_{i-1})$; $\mathbf{I}_{out}=\operatorname{Conv}(\mathbf{F}^{de}_{N})$.
\revise{We applied skip-connections to features generated by SMART layers to capture more informative facial features.}
$(\cdot)\downarrow_r$ and $(\cdot)\uparrow_r$ indicate the bilinear downsampling and upsampling operations with the scale factor $r$. \revise{The $f(\cdot)$ is the feature summation function.}


\begin{figure*}[t]
    \centering
	\includegraphics[width=0.69\textwidth]{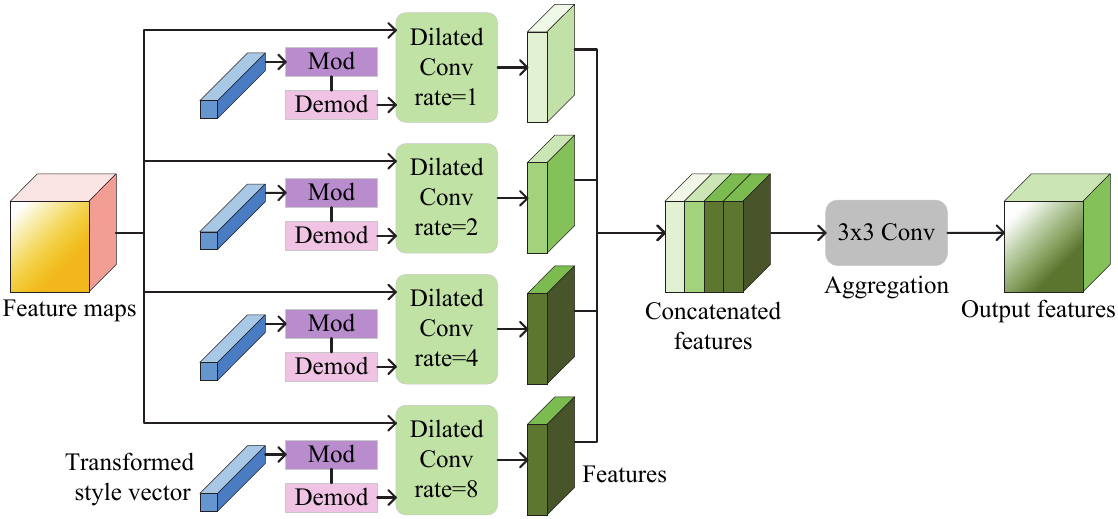}
     \caption{
    Illustration of the style-modulated aggregation transformation (SMART).
    } 
	\label{fig:fig_SMART}
\end{figure*}

\textbf{Style-modulated aggregation transformation.}
As shown in Fig.~\ref{fig:fig_SMART}, the SMART layer uses style prompts and Convolution kernels with various dilation factors to expand the receptive field, allowing the network to capture detailed patterns and distant image contexts more effectively.

Given input feature maps $\mathbf{\bar{F}}$ and corresponding style prompts $\bar{\boldsymbol{w}}$ ($[\boldsymbol{w}^0_{i},\hat{\boldsymbol{z}}]$ for layers in restoration encoder, $[\boldsymbol{w}^0_{i},\hat{\boldsymbol{z}},\boldsymbol{c}]$ for layers in restoration decoder, where $\boldsymbol{w}^0_{i} \in \mathbb{R}^{512 \times 1}$, $\hat{\boldsymbol{z}} \in \mathbb{R}^{512 \times 1}$, $\boldsymbol{c} \in \mathbb{R}^{512 \times 2}$). We first generate a style vector $\boldsymbol{s}  \in \mathbb{R}^{512 \times 1}$ for the subsequent modulation and demodulation~\cite{Karras2020} to rescale Convolution kernels $\boldsymbol{k}$, \textit{e.g.}, $\boldsymbol{k} \in \mathbb{R}^{512 \times 3 \times 3 \times 512}$. The Convolution layers $\operatorname{Conv}^{\boldsymbol{k}'}_{d=m}$, with the reweighted $\boldsymbol{k}'$ and dilation factor $m$  are then applied to extract feature maps to get results. Our SMART layer is expressed as follows:
\begin{equation}\label{equ:SMART_layer}
	\begin{aligned}
	&\boldsymbol{s}= \operatorname{FC}(\bar{\boldsymbol{w}}), \boldsymbol{k}' = \operatorname{Demod}(\operatorname{Mod}(\boldsymbol{k},\boldsymbol{s})),\\
    &\mathbf{F}^d = \phi(\operatorname{Conv}^{\boldsymbol{k}'}_{d}(\mathbf{\bar{F}})), d \in [1,2,4,8],\\
        &\hat{\mathbf{F}} = \operatorname{Aggregation}([\mathbf{F}^1,\mathbf{F}^2,\mathbf{F}^4,\mathbf{F}^8]),\\
	\end{aligned}
\end{equation}
where the modulation operation $\operatorname{Mod}(\cdot)$ is employed to scale the convolution weights using given transformed $\boldsymbol{s}$ style vector, while the demodulation operation $\operatorname{Demod}(\cdot)$ is utilized to rescale kernels back to unit standard deviation for stable training.
{We use a $3 \times 3$ convolution layer as the aggregation layer to merge features from four different dilation rates, enhancing the distinctness of facial structures. Integrating the SMART layer within the restoration network allows our model to capture both global facial structures and local details, while modulated styles provide control over stylistic features.}


\subsection{Module training}\label{sec:co-modulated}
\revise{We use the pre-trained StyleGAN generator $S_{\theta_{s}}$ as our facial feature bank. We first train the style encoder $E_{\theta_{e}}$, followed by optimizing the code diffuser $P_{\theta_{p}}$. Finally, we train the restoration auto-encoder and discriminator.}


\textbf{Style encoder Training.}  
We adopt the style encoder network architecture and training objectives from e4e~\cite{Tov2021}. Throughout the training phase, degraded images are input into the style encoder $E_{\hat{\theta}_{e}}$.

\textbf{Code diffuser training.}  
To learn the denoising process for the generation of denoised latent codes, we reparameterize the learnable Gaussian transition as our code diffuser $P_{\theta_p}(\cdot)$, and use the commonly used optimization objectives, including diffusion loss, Learned Perceptual Image Patch Similarity (LPIPS) loss, and identity loss to constrain the training process.

\textit{Diffusion loss.} The diffusion loss $\mathcal{L}_{dm}$~\cite{NEURIPS2020DDPM} is presented as: $\mathcal{L}_{dm}^{(\theta_{p})}(\boldsymbol{\epsilon},\hat{\boldsymbol{\epsilon}}^{t})= \mathbb{E}_{t,\boldsymbol{w}^0,\epsilon \sim \mathcal{N}(\mathbf{0},\mathbf{I})} \left[ \| \boldsymbol{\epsilon} - \hat{\boldsymbol{\epsilon}}^{t}  \| ^2 \right]$,
where $\hat{\boldsymbol{\epsilon}}^{t} = P_{\theta_{p}}(\boldsymbol{w}^t,\hat{\boldsymbol{w}},t)$. By minimizing $\mathcal{L}_{dm}$, our code diffuser can be trained to predict the added Gaussian noise at step $t$.

\textit{LPIPS loss.} 
To enhance the fidelity of the denoised latent codes, we apply the LPIPS loss~\cite{Zhang2018}  to encourage the perceptual similarity between the inverted image $\mathbf{I}_{ve}$ and the ground-truth $\mathbf{I}_{gt}$ using StyleGAN model $S_{\theta_{s}}$. When $t=T$, we can generate an inverted image $(\mathcal{F}^s,\mathbf{I}_{ve}) = S_{\theta_{s}}(\boldsymbol{w}^0)$ from the denoised latent codes $\boldsymbol{w}^0$. The LPIPS loss is $\mathcal{L}^{(\theta_{p})}_{lpips}(\mathbf{I}_{ve},\mathbf{I}_{gt})= \| \operatorname{VGG}(\mathbf{I}_{ve})-\operatorname{VGG}(\mathbf{I}_{gt})\|_{2}$,
where $\operatorname{VGG}(\cdot)$ is the pre-trained VGG network~\cite{Zhang2018}.


\textit{Identity loss.} 
We apply the identity loss~\cite{Wang_2022_CVPR} to constrain identity similarity between the inverted image $\mathbf{I}_{ve}$ and the ground-truth $\mathbf{I}_{gt}$ in the facial embedding space. It is formulated as: $\mathcal{L}^{(\theta_{p})}_{id}(\mathbf{I}_{ve},\mathbf{I}_{gt})= 1- \cos\left(\operatorname{R}(\mathbf{I}_{ve}),\operatorname{R}(\mathbf{I}_{gt})\right)$.
$\operatorname{R}(\cdot)$ is the pre-trained face recognition ArcFace network.


\textit{Total loss.} 
The total training loss of our code diffuser is expressed as: ${O}(\theta_{p}) = \mathcal{L}_{dm}^{(\theta_{p})} +{\lambda}_{lpips}\mathcal{L}^{(\theta_{p})}_{lpips} +  {\lambda}_{id}\mathcal{L}^{(\theta_{p})}_{id}$,
where ${\lambda}_{lpips}=0.1$ and ${\lambda}_{id}=0.1$ are weights of objectives. During training, we optimize parameters ${\theta}_{p}$ by minimizing the total loss.

\textbf{Restoration networks training.}
We train the restoration and mapping networks using LPIPS loss, identity loss, and adversarial loss.

\textit{LPIPS loss.}\label{sec:LPIPS_loss}
The LPIPS loss  ${\mathcal{L}}^{(\theta_g,\theta_f)}_{lpips}(\mathbf{I}_{out},\mathbf{I}_{gt})$ is applied to enforce perceptual similarity between the restored image $\mathbf{I}_{out}$ and the ground-truth $\mathbf{I}_{gt}$.  

\textit{Identity loss.} The identity loss  ${\mathcal{L}}^{(\theta_g,\theta_f)}_{id}(\mathbf{I}_{out},\mathbf{I}_{gt})$ is employed to constrain identity similarity between the restored $\mathbf{I}_{out}$ and the ground-truth $\mathbf{I}_{gt}$.

\textit{Adversarial loss.}
We employ the adversarial non-saturating logistic loss~\cite{Karras2020}, with $R_{1}$ regularization. The adversarial objective is defined as:
\begin{equation}\label{equ:loss_adv}
\begin{aligned}
\mathcal{L}^{(\theta_{g},\theta_f,\theta_{d})}_{adv}(\mathbf{I}_{out},\mathbf{I}_{gt}) = \mathbb{E}_{\mathbf{I}_{out}}[\log(1-D(\mathbf{I}_{out})] \\+\mathbb{E}_{\mathbf{I}_{gt}}[\log(D(\mathbf{I}_{gt}))] - \frac{\gamma}{2} \mathbb{E}_{\mathbf{I}_{gt}}[\|  \nabla_{\mathbf{I}_{gt}}& D(\mathbf{I}_{gt})\|_{2}^{2}],\\
\end{aligned}
\end{equation}
where $\gamma=10$ is used to balance the $R_1$ regularization term. 
The restoration network $G$ learns to produce realistic images $\mathbf{I}_{out}$, while the discriminator $D$ tries to recognize between real $\mathbf{I}_{gt}$ and restored $\mathbf{I}_{out}$ images. 


\textit{Total objective.} The total training objective is defined as: ${O}(\theta_{g},\theta_{f},\theta_{d}) = \mathcal{L}^{(\theta_{g},\theta_f,\theta_{d})}_{adv}(\mathbf{I}_{out},\mathbf{I}_{gt}) + \hat{\lambda}_{id}\mathcal{L}^{(\theta_{g},\theta_f)}_{id}(\mathbf{I}_{out},\mathbf{I}_{gt}) + \hat{\lambda}_{lpips}\mathcal{L}^{(\theta_{g},\theta_f)}_{lpips}(\mathbf{I}_{out},\mathbf{I}_{gt})$.
We empirically set $\hat{\lambda}_{id}=0.1$ and  $\hat{\lambda}_{lpips}=0.5$ in this work. We can obtain the optimized parameters $({\theta}_{g},{\theta}_{f})$, and ${\theta}_{d}$ via the alternating training phases: $({\theta}_{g}, {\theta}_{f}) =\arg \min _{\theta_{g}, \theta_{f}} {O}(\theta_{g}, \theta_{f}, \theta_{d})$; $({\theta}_{d}) =\arg \max _{\theta_{d}} {O}(\theta_{g}, \theta_{f}, \theta_{d})$.

\begin{table*}[t]
    \centering
    \caption{Quantitative comparison across multiple metrics on the CelebA-Test dataset. \revise{\textbf{Bold} and \underline{Underline} represent the Top-1 and Top-2 performance, respectively.}}
    \label{tab:celeba_test}
    \begin{tabular}{c|c|c|c|c|c|c|c|c||c}
    \hline
    Method & FID$^\dagger$  $\downarrow$ & U-IDS $\uparrow$ & P-IDS $\uparrow$ & FID $\downarrow$ & NIQE $\downarrow$ & \revise{LPIPS $\downarrow$} & PSNR $\uparrow$ & SSIM $\uparrow$ & \revise{Mean rank $\downarrow$}\\ \hline
    PSFRGAN~\cite{Chen_2021_CVPR} & 22.93 & 0 & 0 & 43.88 & 4.27  &  \revise{0.4199} & 24.45 & 0.6308 & \revise{9.75} \\ 
    Wan \textit{et al.}~\cite{Wan_2020_CVPR} & 32.34 & 0 & 0 & 70.21 &  5.19 &  \revise{0.4058}   & 23.00 & 0.6189 & \revise{12.00}\\ 
    PULSE~\cite{Menon_2020_CVPR} & 57.92 & 0 & 0 & 67.75 & 4.71 &  \revise{0.5387}   & 21.61 & 0.6287 & \revise{12.00}\\ 
    GPEN~\cite{Yang_2021_CVPR} & 14.65 & 0.03\% & 0 & 41.99 & 4.25 &   \revise{0.4003}    & {24.63} & 0.6476 &\revise{7.25}\\ 
    GFP-GAN~\cite{Wang_2021_CVPR} & 17.29 & 0 & 0 & 42.39 & 4.58 &   \revise{0.3762}   & 24.46 & {0.6684}&  \revise{7.75} \\ 
    VQFR~\cite{gu2022vqfr} & 13.41 & 0.02\% & 0 & 46.77 & 4.19 &   \revise{0.3557}    & 23.76 & 0.6591 &\revise{6.88} \\ 
    \revise{CodeFormer~\cite{CodeFormer}} & \revise{14.98} & \revise{0} & \revise{0} & \revise{52.42} & \revise{4.74} &   \revise{{0.3432}}    & \revise{{25.14}} & \revise{0.6700} &\revise{7.00} \\ 
    RestoreFormer~\cite{Wang_2022_CVPR} & {12.42} & 0.06\% & 0 & 41.45 & 4.18  &   \revise{0.3654}    & 24.42 & 0.6404 &  \revise{6.25}\\ 
    \revise{RestoreFormer++~\cite{wang2023restoreformer++}} & \revise{{10.95}} & \revise{1.65\%} &\revise{0.23\%} & \revise{40.57} & \revise{4.11} & \revise{{0.3441}}   
     & \revise{{25.31}} & \revise{0.6703} & \revise{\textbf{3.00}} \\
    \revise{DR2~\cite{Wang_2023_CVPR}} & \revise{24.08} & \revise{0} & \revise{0} & \revise{48.44} & \revise{5.89} &  \revise{0.3826} & \revise{{25.47}} & \revise{{0.7066}} &  \revise{7.88}\\ 
    \revise{DiffIR~\cite{Xia_2023_ICCV}}& \revise{19.02} & \revise{0} & \revise{0} &\revise{49.98}  & \revise{4.65}
   &    \revise{0.3582}   & \revise{{26.01}} & \revise{{0.6905}} & \revise{6.50} \\ 
    \revise{PGDiff~\cite{yang2023pgdiff}}  & \revise{14.84} & \revise{0} & \revise{0} & \revise{40.48} & \revise{{3.92}} & \revise{0.4041} & \revise{23.05} & \revise{0.6397} &   \revise{7.75}  \\ 
    \revise{DiffBFR~\cite{Qiu2023DiffBFRBD}}&\revise{11.89}  & \revise{{2.16\%}} & \revise{{0.34\%}} & \revise{38.36}  & \revise{5.49}   &   \revise{0.3685}    & \revise{24.60}  & \revise{0.6594} &  \revise{5.63}\\ 
    \revise{DifFace~\cite{DifFace}} & \revise{13.04} & \revise{0.18\%} & \revise{0} & \revise{{37.88}} & \revise{4.21} & \revise{0.3994}  & \revise{24.80} & \revise{0.6726} & \revise{4.75} \\ 
    \hline
    Ours & {11.76} & {2.21\%} & {0.40\%} & {38.34} & {{4.04}} &  \revise{0.3677}     & {24.60} & {0.6513} & \revise{\underline{3.75}}\\ \hline
    \end{tabular}
\end{table*}

\section{Experimental results and comparisons}\label{sec:experiments}

\subsection{Settings}\label{sec:exp_settings}

\textbf{Implementation.}
The proposed framework was developed using Python and PyTorch.
We fine-tuned the style encoder from e4e~\cite{Tov2021} on our training dataset, with 150,000 iterations using a batch size of 16.
For the code diffuser, we trained 200,000 iterations with a batch size of 8 and a denoising step of 4 $(T=4)$ for both training and testing.
The variance schedule $\beta_t$ increased linearly from $\beta_1 = 0.1$ to $\beta_T = 0.99$. 
The restoration auto-encoder was trained with 500,000 iterations using a batch size of 4. 
Both the code diffuser and restoration auto-encoder used the Adam optimizer (momentum coefficients 0.5 and 0.999, learning rate 0.0002).
\revise{Unless otherwise stated}, all experiments were conducted on the NVIDIA Tesla V100 GPU. 


\textbf{Datasets.}~\label{sec:exp_data}
We trained our algorithm on the FFHQ dataset~\cite{Karras2019}, which comprises 70,000 high-resolution facial images, resized to $512 \times 512$ resolution. 
From the FFHQ dataset, we generated degraded images using the degradation model described in established literature~\cite{Wang_2022_CVPR,Wang_2021_CVPR}. The degradation process is  $\mathbf{I}_{de}=\left\{\left[\left(\mathbf{I}_{gt} \otimes \mathbf{k}_\sigma\right) \downarrow_r+\mathbf{n}_\delta\right]_{J P E G_q}\right\} \uparrow_r$. Here, $\mathbf{I}_{de}$ and $\mathbf{I}_{gt}$ denote a degraded image and its corresponding high-quality counterpart, respectively. $\mathbf{k}_\sigma$ means the Gaussian blur kernel with $\sigma$.  $\mathbf{n}_\delta$ denotes the Gaussian noise with $\delta$. $[\cdot]_{J P E G_q}$ is a JPEG compression with the quality factor $q$. 
During training, the parameters $\sigma$, $r$, $\delta$, and  $q$ were randomly sampled from the given ranges ($\sigma \in [0.2,10]$, $r \in [1,8]$, $\delta \in [0,20]$, and  $q \in [60,100]$ ).

For testing, our method was evaluated on both a synthetic dataset (CelebA-Test~\cite{Liu_2015_ICCV}) and three real-world datasets (LFW-Test~\cite{LFWTech}, CelebChild-Test~\cite{Wang_2021_CVPR}, and WebPhoto-Test~\cite{Wang_2021_CVPR}). The CelebA-Test~\cite{Liu_2015_ICCV} comprises 3,000 samples generated from CelebA-HQ images~\cite{Karras2018} using the above degradation process.
LFW-Test has 1,711 images, which are extracted from the first image of each identity in LFW's~\cite{LFWTech} validation set. 
CelebChild-Test and WebPhoto-Test~\cite{Wang_2021_CVPR} contain 180 and 407 degraded facial images, respectively.
Note that real-world datasets have no ground-truth images.


\textbf{Metrics.}
We employed multiple evaluation metrics including the Fréchet Inception Distance (FID), paired and unpaired inception discriminative scores (P-IDS/U-IDS)~\cite{Zhao2021}, the Naturalness Image Quality Evaluator (NIQE), the Peak Signal-to-Noise Ratio (PSNR), and the Structural Similarity Index Measure (SSIM). 
FID and NIQE have been recognized as benchmark non-reference metrics for assessing the quality of restored images. 
We set the term FID for unpaired evaluation and FID$^\dagger$ for paired evaluation.
\revise{We also ranked each method's performance across all metrics using the mean rank, offering a comprehensive evaluation of overall performance.}

\begin{table*}[t]
\centering
\caption{Quantitative comparisons on three real-world datasets. \revise{\textbf{Bold} and  \underline{Underline} represent the Top-1 and Top-2 performance, respectively.}}
\label{tab:three_real_test}
    \begin{tabular}{c|cc|cc|cc||c}
    \hline
    \multirow{2}{*}{Method}& \multicolumn{2}{c|}{LFW-Test} & \multicolumn{2}{c|}{CelebChild-Test} & \multicolumn{2}{c||}{WebPhoto-Test} &\multirow{2}{*}{\revise{Mean rank $\downarrow$}}\\     \cline{2-7}
     & FID $\downarrow$ & NIQE $\downarrow$ & FID $\downarrow$ & NIQE $\downarrow$ & FID $\downarrow$ & NIQE $\downarrow$ \\ \hline
    PSFRGAN~\cite{Chen_2021_CVPR} & 49.53 & 4.13 & 106.61 & 4.68 & 84.98 & 4.36 & \revise{7.33} \\ 
    Wan \textit{et al.}~\cite{Wan_2020_CVPR} & 60.58 & 5.09 & 117.37 & 5.18 & 101.37 & 5.53 & \revise{13.67}\\ 
    PULSE~\cite{Menon_2020_CVPR} & 64.86 & 4.34 & {102.74} & 4.28 & 86.45 & 4.38 & \revise{8.17} \\ 
    GPEN~\cite{Yang_2021_CVPR} & 51.92 & 4.05 & 107.22 & 4.46 & 80.58 & 4.65 & \revise{8.00} \\ 
    GFP-GAN~\cite{Wang_2021_CVPR} & 49.96 & 4.56 & 111.78 & 4.60 & 87.35 & 5.80 & \revise{10.83}\\ 
    VQFR~\cite{gu2022vqfr} & 49.79 & {3.82} & 114.79 & 4.42 & 84.78 & 4.67 &\revise{7.67} \\ 
    \revise{CodeFormer~\cite{CodeFormer}} &  \revise{52.35} & \revise{4.50} & \revise{116.27} & \revise{5.00}  & \revise{83.17} & \revise{4.86} & \revise{10.83} \\
    RestoreFormer~\cite{Wang_2022_CVPR} & {48.39} & 3.97 & {101.22} & 4.35 & 77.33 & 4.38 & \revise{4.67}\\ 
    \revise{RestoreFormer++~\cite{wang2023restoreformer++}} & \revise{50.25} & \revise{{3.79}} & \revise{105.17} & \revise{{4.15}} & \revise{{75.06}} &\revise{{4.10}} & \revise{\underline{3.83}} \\ 
    \revise{DR2~\cite{Wang_2023_CVPR}} &  \revise{{45.22}} & \revise{4.93} & \revise{120.69} & \revise{4.96}  & \revise{101.57} & \revise{{6.42}} & \revise{11.83} \\
    \revise{DiffIR~\cite{Xia_2023_ICCV}}& \revise{46.80} & \revise{4.31}  & \revise{117.13} & \revise{5.15} &  \revise{85.70} & \revise{5.14} & \revise{10.17} \\ 
    \revise{PGDiff~\cite{yang2023pgdiff}} &  \revise{{40.57}} & \revise{{3.79}} & \revise{112.34} & \revise{{4.02}}  & \revise{90.37} & \revise{{3.92}} & \revise{4.50} \\
    \revise{DiffBFR~\cite{Qiu2023DiffBFRBD}}& \revise{51.26} & \revise{5.47} & \revise{105.84} & \revise{5.52} & \revise{77.89} & \revise{5.71} & \revise{10.50} \\ 
    \revise{DifFace~\cite{DifFace}} &  \revise{{46.31}} & \revise{3.91} & \revise{104.66} & \revise{{4.12}}  & \revise{80.49} & \revise{4.37} & \revise{\underline{3.83}}\\ 
    \hline
    Ours & {46.48} & {{3.85}} & {109.52} & {4.12} & {74.25} &{4.28} & \revise{\textbf{3.67}} \\ \hline
    \end{tabular}
\end{table*}

\begin{figure*}[!h]
    \centering
	\includegraphics[width=0.8\textwidth]{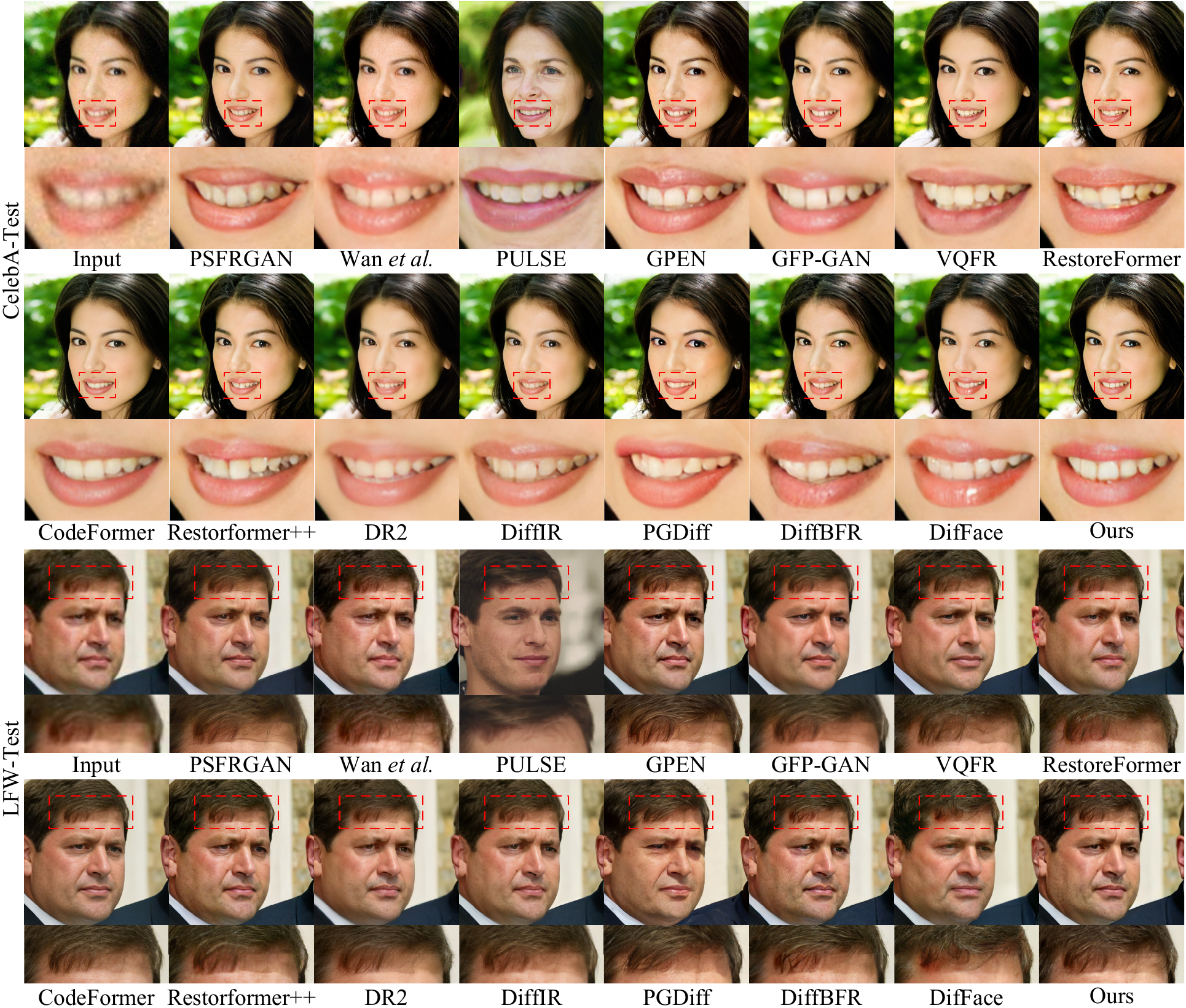}
     \caption{
     Visual comparisons of our method and the SOTA facial restoration methods.
    } 
	\label{fig:fig_qualitative1}
\end{figure*}
\begin{figure*}[!h]
    \includegraphics[width=0.8\textwidth]{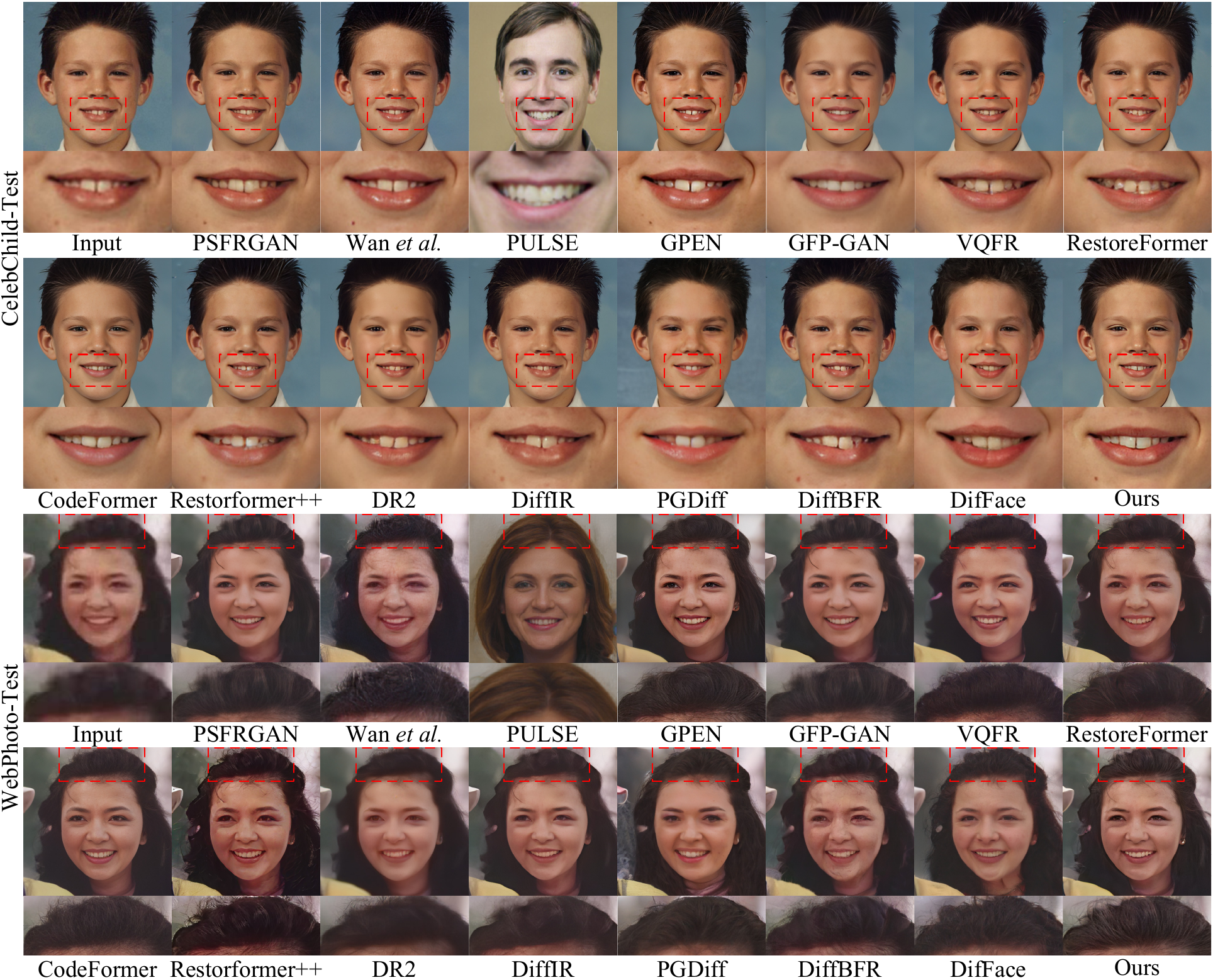}
     \centering
     \caption{
     Visual comparisons of our method and the SOTA facial restoration methods.
    } 
	\label{fig:fig_qualitative2}
\end{figure*}

\subsection{Comparison with SOTA methods}
We compared with SOTA facial restoration methods, including geometric-prior-based (PSFRGAN~\cite{Chen_2021_CVPR}), generative-prior-based (Wan \textit{et al.}~\cite{Wan_2020_CVPR}, PULSE~\cite{Menon_2020_CVPR}, GPEN~\cite{Yang_2021_CVPR}, GFP-GAN~\cite{Wang_2021_CVPR}), codebook-based (VQFR~\cite{gu2022vqfr},  \revise{CodeFormer~\cite{CodeFormer}}, RestoreFormer~\cite{Wang_2022_CVPR}, \revise{RestoreFormer++~\cite{wang2023restoreformer++}}), \revise{and diffusion-based (DR2~\cite{Wang_2023_CVPR}, DiffIR~\cite{Xia_2023_ICCV},  PGDiff~\cite{yang2023pgdiff}, DiffBFR~\cite{Qiu2023DiffBFRBD}, and DifFace~\cite{DifFace}) methods}.

\textbf{Quantitative performance on a synthetic dataset.}
Table~\ref{tab:celeba_test} presents the quantitative results of compared methods on the CelebA-Test dataset. 
\revise{The P-IDS and U-IDS scores measure the percentage of restored images that a trained SVM cannot distinguish from ground-truth images. Most methods scored zero, indicating the SVM easily differentiates between the restored and ground-truth images, suggesting room for improvement in image quality for these methods.}
This table \revise{also} highlights that most generative-prior-based methods outperform geometric-prior-based methods, owing to the extensive facial attributes accessible from generative priors.
Methods like GPEN~\cite{Yang_2021_CVPR} and GFP-GAN~\cite{Wang_2021_CVPR}, which build upon StyleGAN pre-trained priors, achieve commendable results by integrating pre-trained facial features. However, as these methods implicitly predict latent codes, their performance can be impacted when the latent code predictions are inaccurate.
Approaches using reconstruction-oriented dictionaries, such as VQFR~\cite{gu2022vqfr},  \revise{CodeFormer~\cite{CodeFormer}, RestoreFormer~\cite{Wang_2022_CVPR}, and RestoreFormer++~\cite{wang2023restoreformer++}} provide a wealth of high-quality facial features, significantly enhancing restoration quality.
However, the vector quantization process~\cite{VQVAE} may introduce quantization error since continuous data must be mapped to the nearest discrete code. 
That can result in a loss of detail, especially for subtle facial features that are important for a realistic reconstruction.
\revise{Diffusion-based methods like DR2~\cite{Wang_2023_CVPR}, PGDiff~\cite{yang2023pgdiff}, DiffBFR~\cite{Qiu2023DiffBFRBD}, and DifFace~\cite{DifFace} excel in recovering facial structures and details but are hindered by slow inference speeds due to operating in pixel space.
DiffIR~\cite{Xia_2023_ICCV} and our method, on the other hand, perform diffusion and denoising in the embedding space of pre-trained models to generate high-quality visual prompts. 
Our approach leverages latent representations from generative models, providing a more intuitive way to guide the face restoration process. 
It shows superior performance in terms of perceptual quality (FID, U-IDS, P-IDS, NIQE) and achieves competitive performance in terms of fidelity (LPIPS, PSNR, SSIM).
We achieved the second-best mean rank (3.75), which is close to the best (3.00), highlighting ours and RestoreFormer++'s superior restoration quality across the eight metrics.
}

\begin{table*}[t] \small
	\centering
	\caption{\revise{Network complexity and computational efficiency of compared methods.}}
 
    \revise{
    	\begin{tabular}{c|cccccccc}
        	\hline 
        	{Method}        &    PSFRGAN &  Wan~\textit{et al.}   &   PULSE  & GPEN &  GFP-GAN& VQFR & CodeFormer   \\
            \hline
            {\# Parameters (M)} &    45.69     &    92.06         &   {26.21}     &   26.23    &     60.76      &     79.30     & \revise{84.22}  \\ 
            {FLOPs (G)} &    411.19     &      545.45       &     {13.28}    &      {17.78}  &      {85.04}    &    1070.00      &  \revise{401.40} \\         
            {FPS} &    5.87    &     21.03        &   0.72    &     {50.76} &     28.72    &    12.01     &   \revise{20.44}    \\ 
           \hline 
           {Method}        & RestoreFormer  & RestoreFormer++  &   DR2  & PGDiff &  DifFace &   \revise{DiffIR}    & Ours   \\
            \hline
            {\# Parameters (M)} &   72.37      &      73.16       &    85.79    &    159.70     &      159.70      &       26.55   &  420.88 \\ 
            {FLOPs (G)} &    340.80     &     341.60        &    725.08    &    185.95    &      185.95     &       224.92   &  570.21  \\ 
            {FPS} &    21.24     &     20.76        &   1.48     &   0.01     &      0.13     &     24.90     &   12.30   \\ 
           \hline 
        \end{tabular}
        }
	\label{tab:model_complexity}
\end{table*}

\textbf{Quantitative performance on real-world datasets.}
Table~\ref{tab:three_real_test} presents evaluations of methods on LFW-Test, CelebChild-Test, and WebPhoto-Test datasets to compare their generalization capabilities.
Methods such as Wan \textit{et al.}~\cite{Wan_2020_CVPR} and PULSE~\cite{Menon_2020_CVPR}, which do not incorporate identity information from degraded images, may experience a drop in performance. GPEN~\cite{Yang_2021_CVPR} and GFP-GAN~\cite{Wang_2021_CVPR} apply generative facial priors, yet their local feature fusion might constrain their effectiveness.
Approaches like VQFR~\cite{gu2022vqfr}, \revise{CodeFormer~\cite{CodeFormer}, RestoreFormer~\cite{Wang_2022_CVPR}, and RestoreFormer++~\cite{wang2023restoreformer++}} rely on a fixed codebook, which may restrict their ability to capture the fine-grained facial attributes.
\revise{Diffusion-based methods, such as PGDiff~\cite{yang2023pgdiff} and DifFace~\cite{DifFace} have shown remarkable performance on three real-world datasets. 
However, they require multiple denoising steps in the pixel space, which would increase computational resource demands. 
In contrast, DiffIR~\cite{Xia_2023_ICCV} and our method perform diffusion and denoising in the embedding space of pre-trained models, achieving an effective balance between restoration quality and inference speed.
Our method achieves the best mean rank (3.67), highlighting our superior perceptual quality across the three real-world datasets.}

\begin{figure*}[!htbp]
	\includegraphics[width=0.8\textwidth]{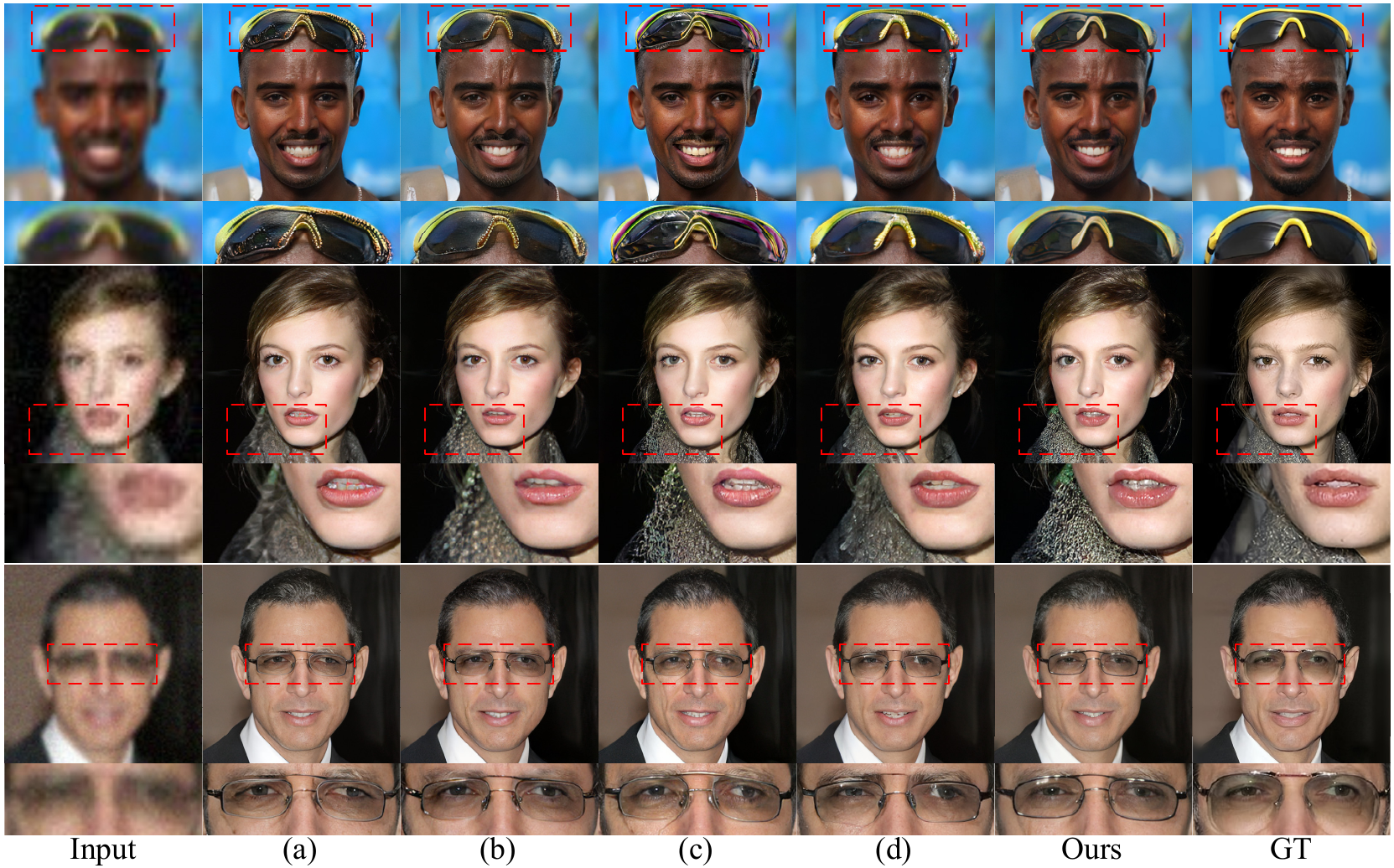}
    \centering
     \caption{
     Visual examples from the ablation study with (a) removing the coder diffuser, (b) replacing our SMART layers with style layers, (c) removing the modulation of style codes, (d) removing facial priors in fusion processes, and the full configurations (ours).
    } 
	\label{fig:fig_ablation_visual_res}
\end{figure*}

\textbf{Qualitative performance.}
Fig.~\ref{fig:fig_qualitative1} and Fig.~\ref{fig:fig_qualitative2} present the visual comparisons across four datasets. 
While most existing approaches can produce high-fidelity faces from degraded images, our method attains more detail in critical regions, such as the hair and mouth.
The results indicate that methods by Wan \textit{et al.}~\cite{Wan_2020_CVPR} sometimes struggle to preserve the identity presented in the original images. GPEN~\cite{Yang_2021_CVPR} and GFP-GAN~\cite{Wang_2021_CVPR} demonstrate improved performance, yet they occasionally introduce visual artifacts on facial features.
VQFR~\cite{gu2022vqfr}, \revise{CodeFormer~\cite{CodeFormer}, RestoreFormer~\cite{Wang_2022_CVPR}, and RestoreFormer++~\cite{wang2023restoreformer++}} exhibit clearer facial features. 
\revise{DR2~\cite{Wang_2023_CVPR}, DiffIR~\cite{Xia_2023_ICCV}, PGDiff~\cite{yang2023pgdiff}, DiffBFR~\cite{Qiu2023DiffBFRBD}, and DifFace~\cite{DifFace} showcase strong capabilities in high-quality facial restoration.}
In contrast, our approach excels in restoring facial details with clearer facial features and fewer artifacts. \revise{For example, our restored teeth maintain a clear structure, free from adhesion or artifacts. 
This is achieved through our visual style prompt learning, which accurately predicts style prompts from degraded images. 
These prompts are used by the facial feature bank to generate candidate features, including detailed structures like teeth. Additionally, our SMART layer enhances reasoning by capturing informative distant image contexts from these candidate features.}



\revise{
\textbf{Analysis on network complexity and computational efficiency.}
Table~\ref{tab:model_complexity} presents the model complexity and computational efficiency of the compared methods, measured in terms of parameters, FLOPs (with one-time inference), and frames per second (FPS) during inference. 
All methods were tested on a single NVIDIA GeForce RTX 3080Ti GPU using $512 \times 512$ images. 
The results indicate that diffusion-based methods generally have slower inference speeds compared to GAN-based and codebook-based methods. 
For example, PGDiff~\cite{yang2023pgdiff} takes approximately 92 seconds per image, which is about 1,135 times longer than our processing time (81 milliseconds).
Models like GPEN, RestoreFormer, and our approach can process over 10 images per second.
GPEN processes 50.76 images per second, demonstrating strong computational efficiency.
Although our model has higher parameters, this large capacity is essential for achieving competitive results. 
Our method achieves a good balance between restoration quality and inference speed.
}


\begin{table}[t]
\centering
\caption{Ablation study of the framework components on three real-world datasets: (a) removing our coder diffuser, (b) replacing our SMART layers with style layers, (c) removing the modulation of style codes, (d) removing facial priors in fusion processes, and the full configurations (ours). \textbf{Bold}  values indicate the best results.}
\label{tab:ablation_real_test}
\resizebox{0.49\textwidth}{!}{
    \begin{tabular}{c|cc|cc|cc}
    \hline
    \multirow{2}{*}{Method}& \multicolumn{2}{c|}{LFW-Test} & \multicolumn{2}{c|}{CelebChild-Test} & \multicolumn{2}{c}{WebPhoto-Test} \\     \cline{2-7}
     & FID $\downarrow$ & NIQE $\downarrow$ & FID $\downarrow$ & NIQE $\downarrow$ & FID $\downarrow$ & NIQE $\downarrow$ \\ \hline
    (a) & 46.58 & 4.13 & 111.37 & 4.36 & 79.20 & 4.51 \\ 
    (b) &  48.75 & 4.09   & 112.19 & 4.21   & 76.34  &  \textbf{4.33}   \\ 
    (c) &46.65 & 4.05 & 111.75 & \textbf{4.19} & 79.71 & 4.43 \\ 
    (d) &49.35 & 3.98 & 110.76 & 4.31 &75.60 & 4.37 \\ 
    \hline
    Ours & \textbf{46.55} & \textbf{3.96} & \textbf{110.09} & \textbf{4.19} & \textbf{74.80} & \textbf{4.33} \\ \hline
    \end{tabular}
    }
\end{table}

\begin{figure*}[t]
	\includegraphics[width=0.80\textwidth]{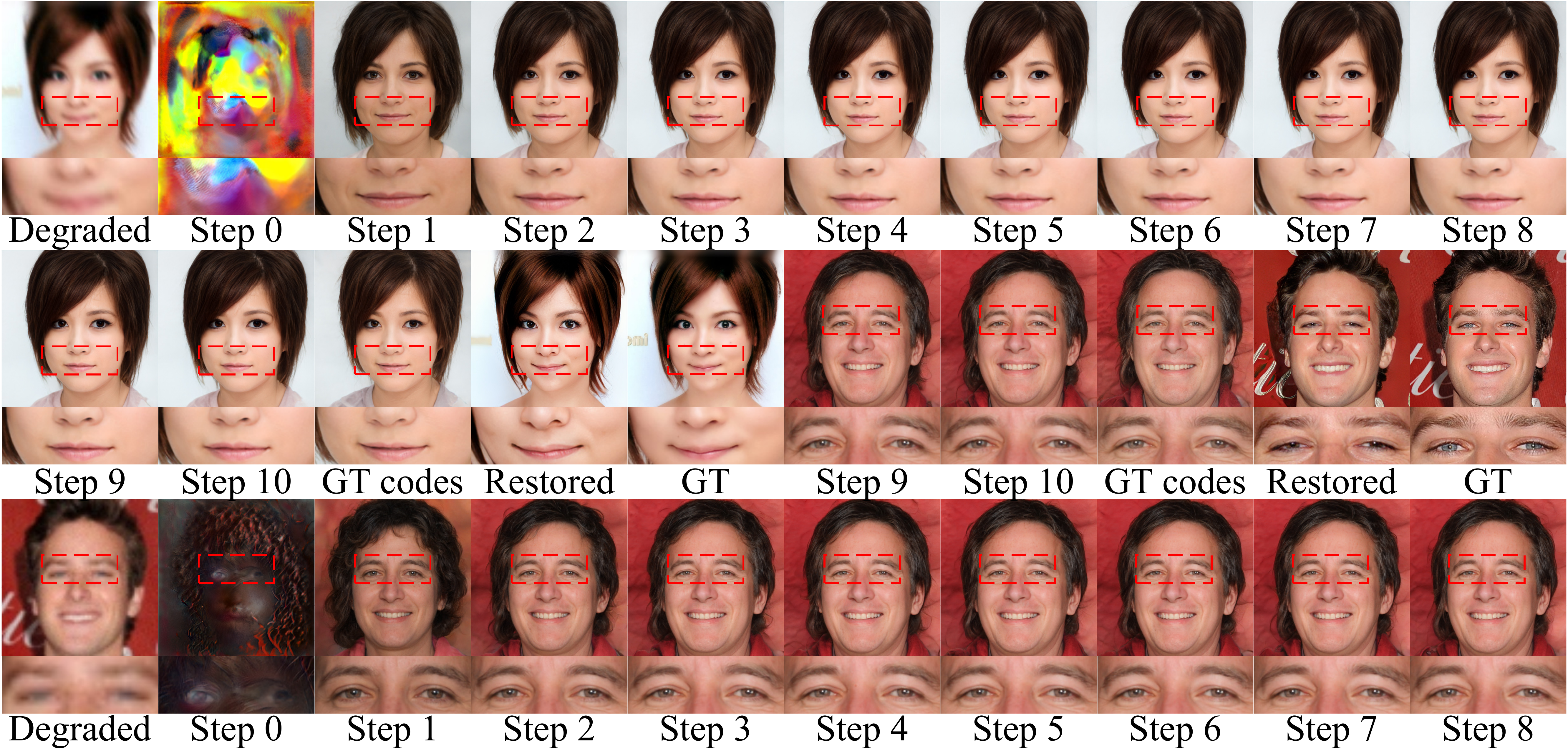}
    \centering
     \caption{
     Visualization of denoised latent codes for each step and restored images. Through the denoising process, the codes incrementally reveal more clear facial features. 
    } 
	\label{fig:fig_code_diffuser_analysis}
\end{figure*}

\begin{figure*}[t]
    \centering
	\includegraphics[width=0.8\textwidth]{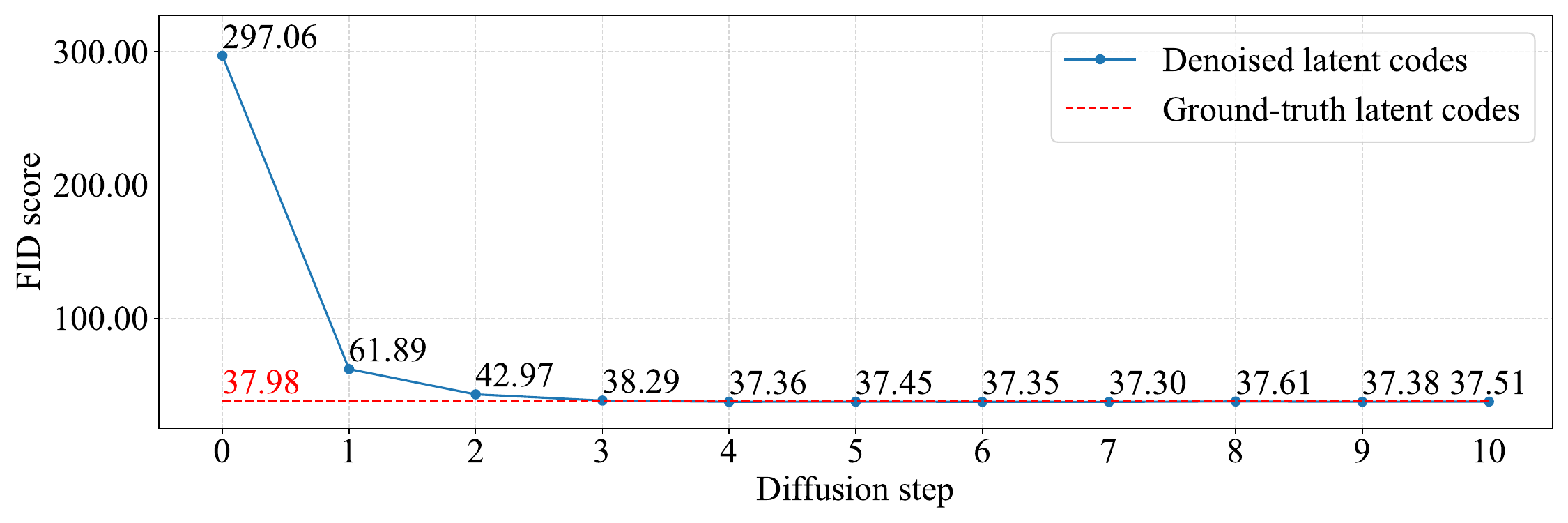}
     \caption{
      FID comparisons between denoised latent codes and ground-truth latent codes.
    } 
	\label{fig:fig_denoised_fid_plot}
\end{figure*}

\subsection{Ablation study}


\revise{We conducted the ablation study to evaluate the impact of individual components within our framework by selectively omitting each one and evaluating its effect on restoration performance. 
Since all models converged well within 400,000 iterations, we used this number for the ablation study, maintaining the same training settings as described in Subsection~\ref{sec:exp_settings}.}


Table~\ref{tab:ablation_real_test} presents the quantitative outcomes for various configurations. 
Our fully-equipped model outperforms alternative variants in FID and NIQE metrics, demonstrating its good restoration capabilities. 
When removing the coder diffuser (a), the restoration network relies solely on the initial codes as the style prompts, which may not accurately capture clean facial attributes from degraded images, leading to a noticeable drop in metric scores.
Replacing the SMART layer with a style layer (b) limits the network's ability to dynamically capture the global facial structure and local details, resulting in worse FID and NIQE performance.
Removing style code modulation (c) restricts the model's ability to leverage style prompts effectively. That leads to suboptimal performance, highlighting the importance of style prompt modulation.
Omitting facial priors in the fusion process (d) degrades the fusion mechanism to a basic Unet skip connection, with the resultant decline in FID and NIQE scores. 
This emphasizes the importance of facial priors for achieving better performance.
Overall, our model excels by integrating finely-tuned style prompts via our coder diffuser, adeptly capturing both local and global facial features with our SMART layer, and enhancing visual quality through the effective use of spatial facial features and facial priors.

%

\begin{figure*}[t]
    \centering
	\includegraphics[width=0.68\textwidth]{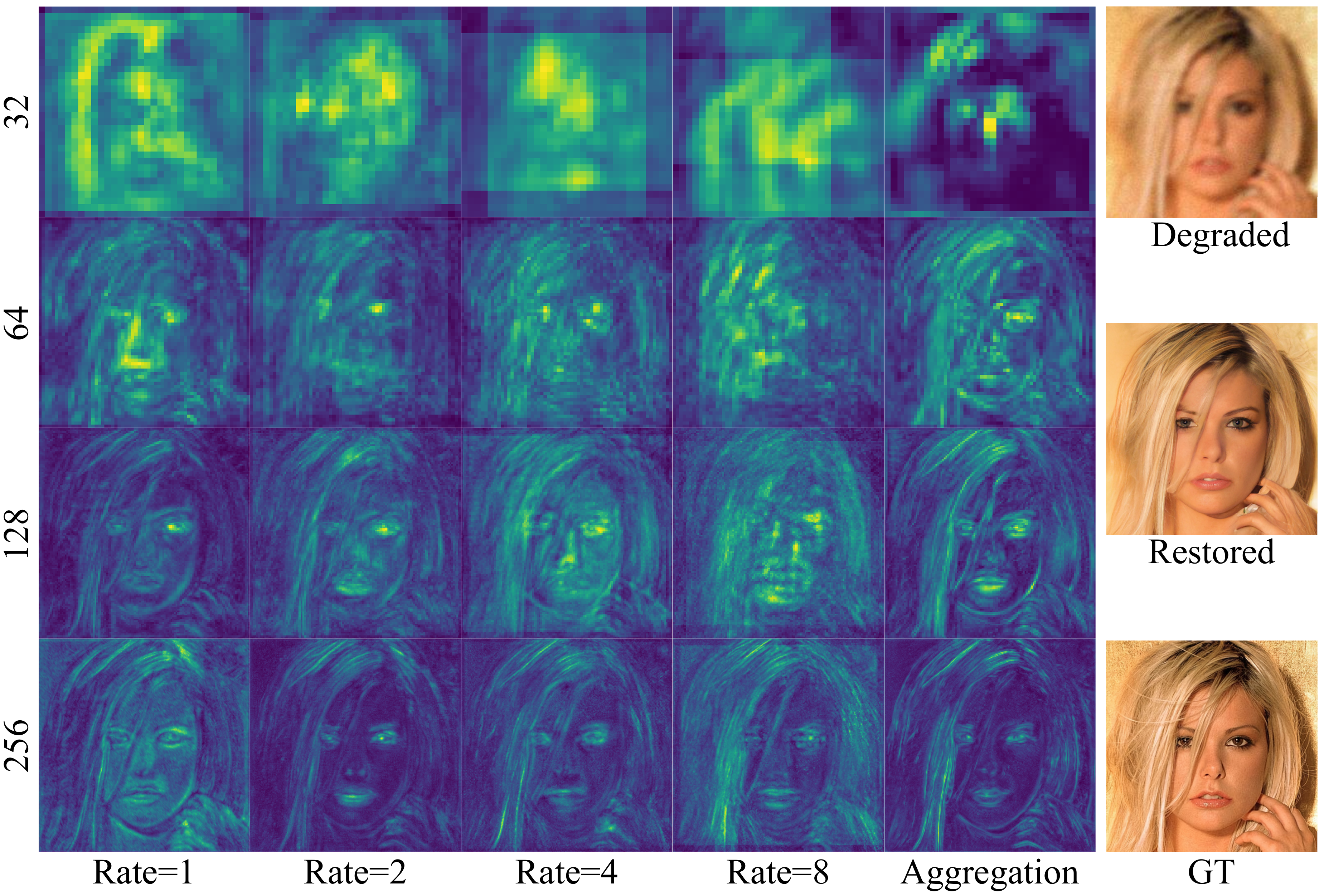}
     \caption{
     Visualization of feature maps from SMART layers at resolutions from $32^2$ to $256^2$, showcasing results across different dilation rates (1, 2, 4, 8) and aggregation layers.
    } 
	\label{fig:fig_SMART_analysis}
\end{figure*}


Fig.~\ref{fig:fig_ablation_visual_res} shows the visual results for each model variant. Removing the coder diffuser (a) results in noticeable artifacts, particularly around the mouth and decorative features. Replacing the SMART layers with style layers (b) slightly blurs the facial structures and details, such as the cloth in the second row, likely due to the style layer's limited receptive field. Omitting style code modulation (c) diminishes image quality, leading to color inconsistencies. In the first row, model (c) exhibits color discrepancies between the hair and sunglasses due to the lack of consistent style modulation.
Removing facial priors from the fusion process (d) impacts structure fidelity, with the third row showing asymmetrical glasses.
In contrast, our full model effectively recovers structures and details, showing superior restoration capabilities.




\subsection{Analysis of style prompt learning}
\revise{We assessed the quality of denoised latent codes by feeding them into the pre-trained StyleGAN generator to produce inverted images on the CelebA-Test dataset. Ground-truth latent codes were obtained by feeding ground-truth images into the style encoder. 
Fig.~\ref{fig:fig_code_diffuser_analysis} visualizes these latent codes alongside degraded, restored, and ground-truth images for analysis.
With more diffusion steps, the denoised latent codes increasingly reflect more distinct facial features and exhibit higher fidelity.
However, beyond 4 denoising steps, further refinement yields similar visual performance. 
Attributes such as facial pose, expression, and hairstyle were closely aligned with the ground-truth latent codes, underscoring the effectiveness of visual style prompt learning in generating consistent coarse facial features. Using our restoration auto-encoder, the restored images closely resemble the ground-truth images in quality.
Table~\ref{tab:ablation_real_test} (a) and Fig.~\ref{fig:fig_code_diffuser_analysis} show the effectiveness of our diffusion-based style prompt module in improving restoration.}




\revise{Fig.~\ref{fig:fig_denoised_fid_plot} shows FID scores for inverted images, demonstrating consistent improvements in latent codes with more diffusion steps, but saturating at step 7. After step 4, the denoised codes even surpass the ground-truth FID score (37.98). After step 7, the performance fluctuates around a score of 37.30, showing our code diffuser's ability to produce high-quality style prompts.}

\subsection{Analysis of SMART layer}

Fig.~\ref{fig:fig_SMART_analysis} visualizes the impact of the SMART layer across varying resolutions during restoration.
Feature maps were generated at dilation rates of 1, 2, 4, and 8, alongside their aggregation layers, spanning resolutions from $32^2$ to $256^2$.
A dilation rate of 1 converts the layer back to an original style layer. 
Lower dilation rates (1 and 2) typically emphasize local features, while higher rates focus on global attributes.
The aggregation process further prioritizes more informative and detailed facial features while reducing the impact of less informative features.
For example, at a $64^2$ resolution, a dilation rate of 1 mainly focuses on the nose and eye regions, while a dilation rate of 8 captures the global facial features, but with some noise. The aggregation of these features results in clearer facial structures. 
Table~\ref{tab:ablation_real_test} (b) and visual results confirm the SMART layer's effectiveness in producing key facial features.

\begin{table}[t]
\centering
\caption{Quantitative comparisons of facial landmark detection in detection success rate (DSR) and normalized mean error (NME). \textbf{Bold} values indicate the best results.}
\label{tab:application_detection}
    \begin{tabular}{c|c|c}
    \hline
      & \multicolumn{1}{c|}{FAN} & \multicolumn{1}{c}{Our improved} \\ \hline
    \multicolumn{1}{c|}{DSR\% $\uparrow$}  &  99.73\%  & \textbf{100.00\%}  \\ \hline
     \multicolumn{1}{c|}{NME\% $\downarrow$}  &  6.08\%  & \textbf{2.43\%} \\ \hline
    \end{tabular}
\end{table}


\begin{figure*}[t]
    \centering
	\includegraphics[width=0.8\textwidth]{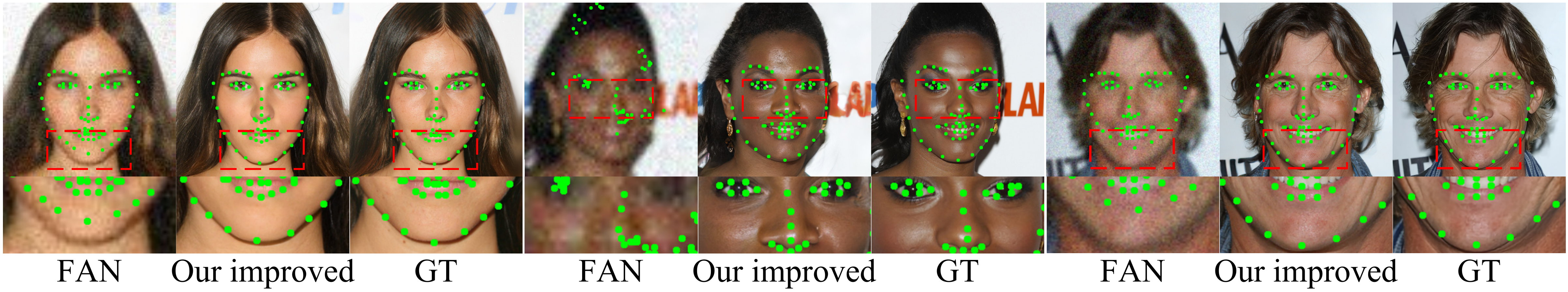}
     \caption{
     Visualization of facial landmark detection on degraded images, restored images, and ground-truth images. Green points correspond to visualized facial landmarks.
    } 
	\label{fig:fig_landmark}
\end{figure*}

\subsection{Applications}
We show our method's effectiveness in improving facial landmark detection and emotion recognition. We tested on the CelebA-Test dataset.


\textbf{Facial landmark detection.}
We integrated our method with a face alignment network (FAN~\cite{Bulat2017}) to develop an improved detection algorithm (Ours + FAN).
We set the degraded images as inputs. The landmarks detected on the ground-truth images were designated as ground-truth landmarks. We recorded the detection success rate (DSR\%) for the face images and the normalized mean error (NME\%) for landmarks.
Table~\ref{tab:application_detection} shows corresponding quantitative comparisons.  
Incorporating our facial restoration method significantly boosts the DSR to 100\%, and reduces the NME score, achieving a relative improvement of 3.65\%. 
This indicates that our approach markedly enhances the detection accuracy by restoring clearer facial features.

Fig.~\ref{fig:fig_landmark} further visualizes the detection comparisons, showing that direct detection on degraded images often leads to inaccurate results. 
Incorporating our restoration method significantly improves facial landmark detection, by enhancing the clarity of facial features in the restored images.

\begin{figure*}[!t]
    \centering
	\includegraphics[width=0.8\textwidth]{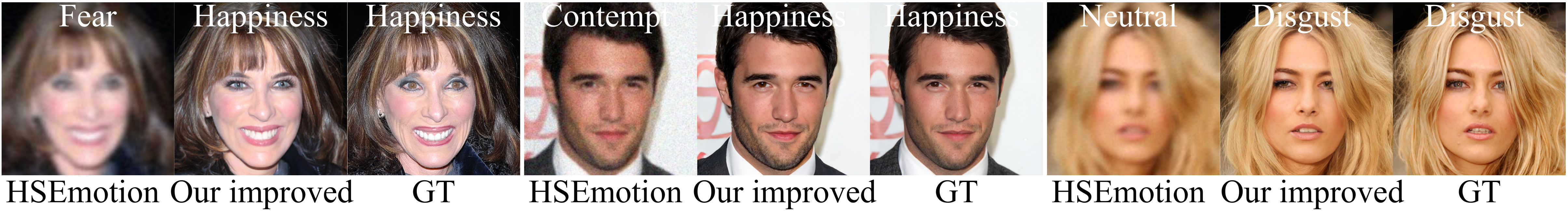}
     \caption{Visualization of face emotion recognition using HSEmotion and our enhanced. 
    } 
	\label{fig:fig_emo_rec}
\end{figure*}


\textbf{Face emotion recognition.}
We incorporated our face restoration into a face emotion recognition algorithm (HSEmotion)~\cite{FER_2023_ICML} to build an enhanced algorithm (Ours + HSEmotion). 
We set the degraded images as inputs and set the emotions recognized from the ground-truth images as the ground-truth labels. 
HSEmotion works well in most cases, achieving 79.73\% recognition accuracy.
Our enhanced one achieved a higher accuracy of 86.03\%, marking a relative improvement of 6.30\% over HSEmotion. 

%

Fig.~\ref{fig:fig_emo_rec} shows visual comparisons for emotion recognition. 
Even the SOTA method struggles with degraded images.
However, by integrating our facial restoration, the enhanced algorithm accurately classifies expressions, highlighting the importance of our restoration in improving recognition accuracy.



\section{Conclusion}\label{sec:conclusion}
In this paper, we have explored a novel visual style prompt learning framework for blind face restoration.
Our approach utilizes a diffusion-based style prompt module to predict high-quality visual cues in the latent space of a pre-trained model. 
A code diffuser aids in the denoising process, while the SMART layer enhances context and detailed feature extraction. 
Experiments demonstrate that our method effectively predicts prompts closely aligned with ground-truth latent representations, leading to superior image restoration. 
Additionally, our technique improves performance in face landmark detection and emotion recognition. 
Ablation studies further validated the effectiveness of our proposed components.

\textbf{Limitations and future work.} \revise{While our method achieves high-quality blind facial restoration, it has limitations. Like most blind facial restoration algorithms, our model may not work well on highly complex backgrounds. That might be caused by the lack of sufficiently diverse background data in the training set. 
Developing a more advanced technique to separate foreground and background restoration could address this challenge.}

\revise{
Our visual prompts correspond to specific facial attributes represented by style latent codes. 
While these prompts are not directly linked to specific textual information, they provide a strong connection between text and facial attributes. 
For example, FFCLIP~\cite{zhu2022one} demonstrated how text prompts can manipulate style latent codes for targeted attribute manipulations, suggesting a potential pathway for integrating textual information with our visual prompts.
In the future, we would like to explore methods to bridge textual prompts with facial features, and adapt our method for broader applications, such as real-world video face restoration~\cite{Chen_2024_CVPR}}.

\bibliographystyle{elsarticle-num} 
\bibliography{main_bib}




\end{document}